\documentclass{article}

\usepackage{microtype}
\usepackage{graphicx}
\usepackage{subfigure}
\usepackage{multirow}
\usepackage{booktabs} 

\usepackage{hyperref}
\usepackage{caption}

\usepackage{xspace}
\usepackage{xcolor}
\usepackage{CJKutf8}

\usepackage[accepted]{icml2021}

\icmltitlerunning{Fused Acoustic and Text Encoding for Multimodal Bilingual Pretraining and Speech Translation}

\usepackage{dsfont}
\usepackage{amsmath}
\usepackage{amssymb}
\usepackage{amsthm}
\usepackage{bm} 

\usepackage[utf8]{inputenc}

\newcommand{\notes}[1]{}

\theoremstyle{definition}

\theoremstyle{plain}

\newcommand{\vech}{\ensuremath{\mathbf{h}}}

\newcommand{\vecs}{\mathbf{s}\xspace}

\newcommand{\ith}[1]{\ensuremath{i^{{th}}}}

\newcount\permx
\newcount\permy
\def\permdot#1#2{
\permx=#1 \advance\permx by-1
\permy=#2 \advance\permy by-1
\psframe[fillcolor=black, fillstyle=solid]
(\permx,\permy)(#1, #2)
}

\newcommand{\boxnum}[1]{{\setlength{\fboxsep}{1pt}\raisebox{1pt}{\hspace{1pt}\fbox{\tiny #1}\hspace{1pt}}}}
\newcommand{\ind}[1]{\ensuremath{_{\kern-0.5pt\boxnum{#1}}}}

\newcommand{\vecx}{\mathbf{x}\xspace}
\newcommand{\vecy}{\mathbf{y}\xspace}
\newcommand{\vece}{\ensuremath{\bm{e}}\xspace}

\newcommand{\smallnt}[1]{\ensuremath{_{\mbox{\tiny PP}}}\xspace}

\newcommand{\pseudocode}{Algorithm}
\floatname{algorithm}{\pseudocode}

\iffalse

\else

\fi

\newcommand{\eos}{\mbox{\scriptsize \texttt{<eos>}}\xspace}

\newcommand{\dsxy}{D_{\vecs,\vecx,\vecy}\xspace}
\newcommand{\dsy}{D_{\vecs,\vecy}\xspace}
\newcommand{\dsx}{D_{\vecs,\vecx}\xspace}
\newcommand{\dxy}{D_{\vecx,\vecy}\xspace}
\newcommand{\ds}{D_{\vecs}\xspace}
\newcommand{\dx}{D_{\vecx}\xspace}
\newcommand{\dy}{D_{\vecy}\xspace}
\begin{document}

\begin{CJK}{UTF8}{gbsn}

\twocolumn[
\icmltitle{
Fused Acoustic and Text Encoding for \\ Multimodal Bilingual
Pretraining and Speech Translation
           }



\icmlsetsymbol{equal}{*}

\begin{icmlauthorlist}
\icmlauthor{Renjie Zheng}{equal,baidu}
\icmlauthor{Junkun Chen}{equal,osu}
\icmlauthor{Mingbo Ma}{baidu}
\icmlauthor{Liang Huang}{baidu,osu}
\end{icmlauthorlist}

\icmlaffiliation{baidu}{Baidu Research, Sunnyvale, CA, USA}
\icmlaffiliation{osu}{Oregon State University, Corvallis, OR, USA}

\icmlcorrespondingauthor{Renjie Zheng}{renjiezheng@baidu.com}

\icmlkeywords{Machine Learning, ICML}

\vskip 0.3in
]



\printAffiliationsAndNotice{\icmlEqualContribution} 

\begin{abstract}
Recently,  representation learning for text and speech
has successfully improved many language related tasks.
However, all existing methods suffer from two limitations:
(a) they only learn from one input modality, 
while a unified representation for both speech and text
is needed by tasks such as end-to-end speech translation,
and as a result, 
(b) they can not exploit various large-scale text and speech data 
and their performance is limited by the scarcity of 
parallel speech translation data.
To address these problems, we propose a Fused Acoustic and
Text Masked Language Model (FAT-MLM) which 
jointly learns a unified representation for 
both acoustic and text input
from various types of corpora including parallel data for speech recognition and 
machine translation, and even pure speech and text data.
Within this crossmodal representation learning framework, we further present an 
end-to-end  model for
Fused Acoustic and Text Speech Translation (FAT-ST).
Experiments on three translation directions show that 
by fine-tuning from FAT-MLM,
our proposed speech translation models
substantially improve  translation quality  by up to $+5.9$ BLEU.
\end{abstract}

\section{Introduction}


In recent years, task-agnostic text representation learning
\cite{peters2018deep, BERT, sun2019ernie}
has attracted much attention in the NLP community due to its
strong performance to many downstream tasks.
More recently, unsupervised speech representation learning
\cite{baevski2020wav2vec,chen2020mam, liu2020tera}
also 
successfully improved many speech related tasks,
such as speech recognition and speech translation.

\begin{figure}[t!]
\centering
\begin{tabular}{l}
\begin{minipage}[t]{1.0 \linewidth}
\begin{center}
\includegraphics[width=4.5cm]{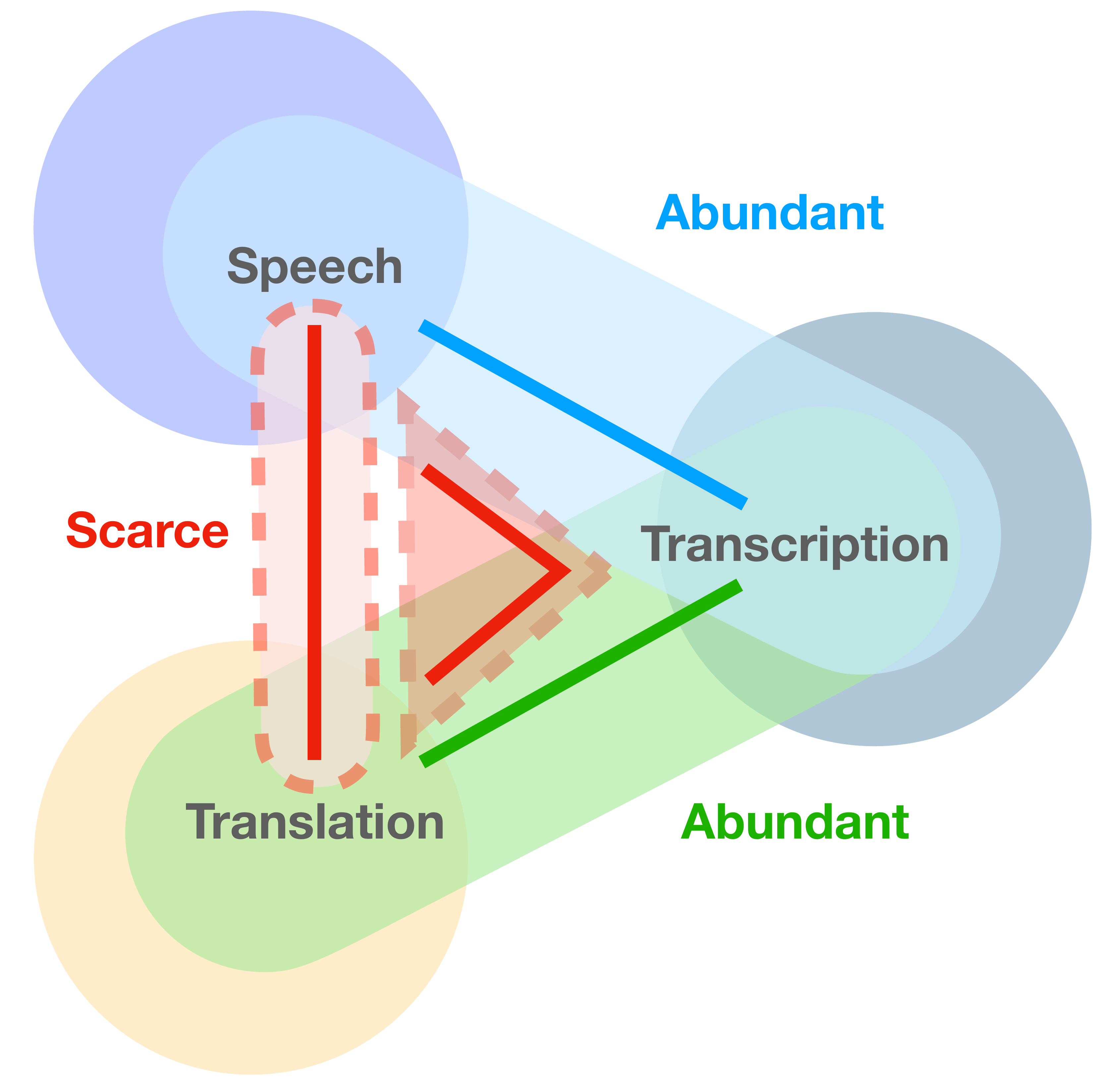}
\end{center}
\end{minipage}
\\
\end{tabular}
\vspace{-0.3cm}
\caption{
The quality of end-to-end speech translation models
has been limited by the scarcity of speech translation datasets. 
However, there is an abundance of datasets for speech, text, speech recognition, and machine translation data that can be leveraged. 
}
\label{fig:intro}
\end{figure}

However all these existing methods can only handle 
one modality, either text or speech, 
while joint acoustic and text representation is desired
for many end-to-end spoken language processing tasks,
such as 
spoken question answering \cite{chuang2019speechbert}
and
end-to-end speech-to-text translation \cite{liu2020bridging}.
For example, end-to-end speech
translation (ST) is desired 
due to its advantages over the pipeline paradigm, 
such as low latency, alleviation of error propagation, and fewer parameters
\cite{weiss2017sequence, berard2018end, jia2019leveraging, sperber2017neural, zheng2020fluent, chen2021direct}.
However, its translation quality
is limited by 
the scarcity of large-scale parallel speech translation data
while
there exists sufficient data for 
speech recognition and text machine translation (Fig.~\ref{fig:intro}).
It would be helpful if 
source speech and bilingual
text can be encoded into a unified representation via
abundant speech recognition and text machine translation data.
\citet{liu2020bridging} show that jointly training
a multi-modal
ST encoder can largely improve the translation quality. 
However, their proposed representation learning method 
is constrained to the sequence-to-sequence framework and 
there is no experiment showing whether their proposed
method can benefit from extra
speech recognition and machine translation
data.

\begin{figure*}[ht!]
\centering
\begin{tabular}{l}
\begin{minipage}[b]{0.5 \linewidth}
\begin{center}
\subfigure[Masked Language Model (MLM) for text  representation learning.
]{
\makebox[\linewidth][c]{
\includegraphics[width=5cm]{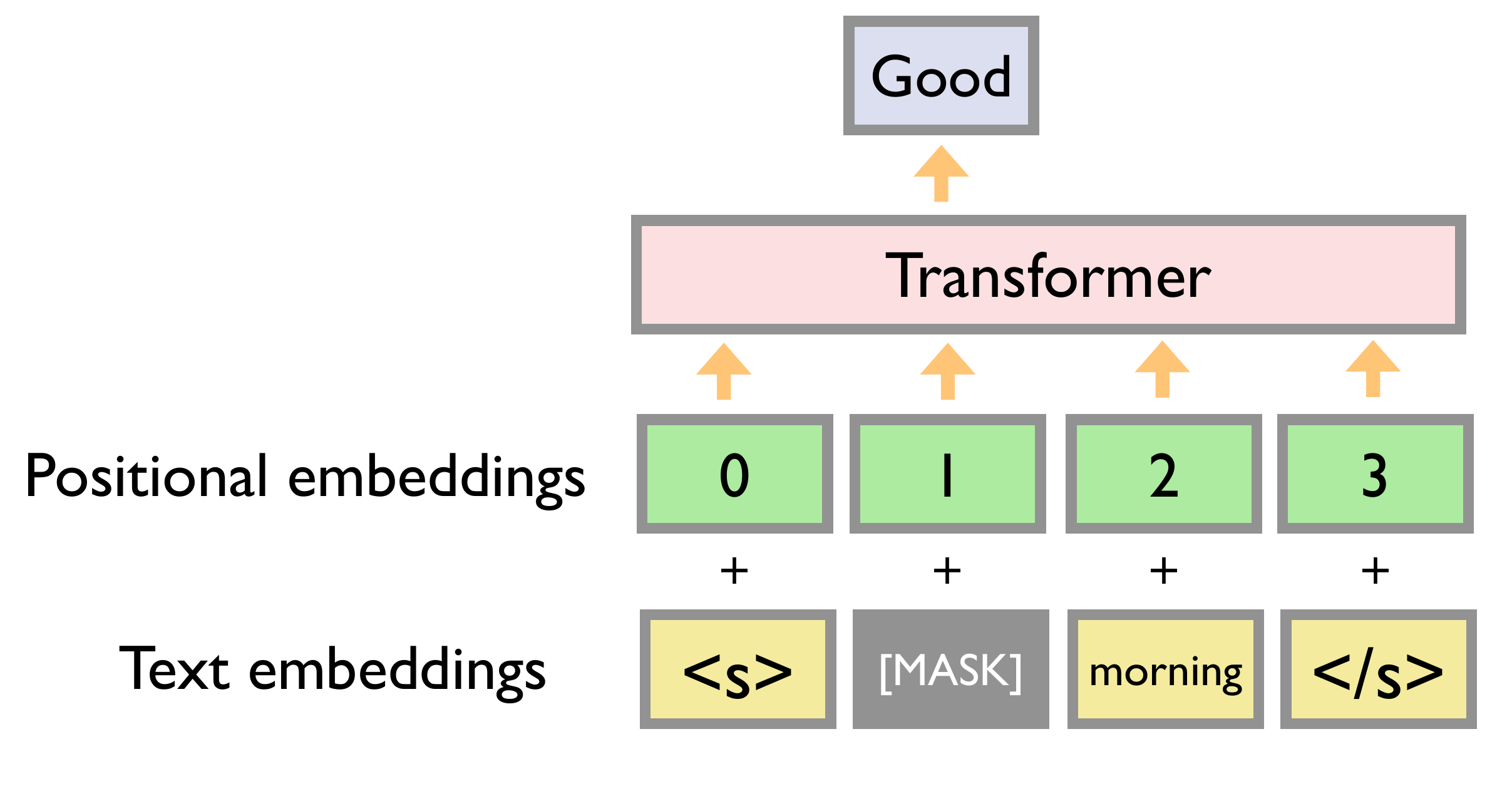}
    }
\label{fig:mlm}

}
\\
\subfigure[Translation Language Model (TLM)
for crosslingual text.]{
\includegraphics[width=8cm]{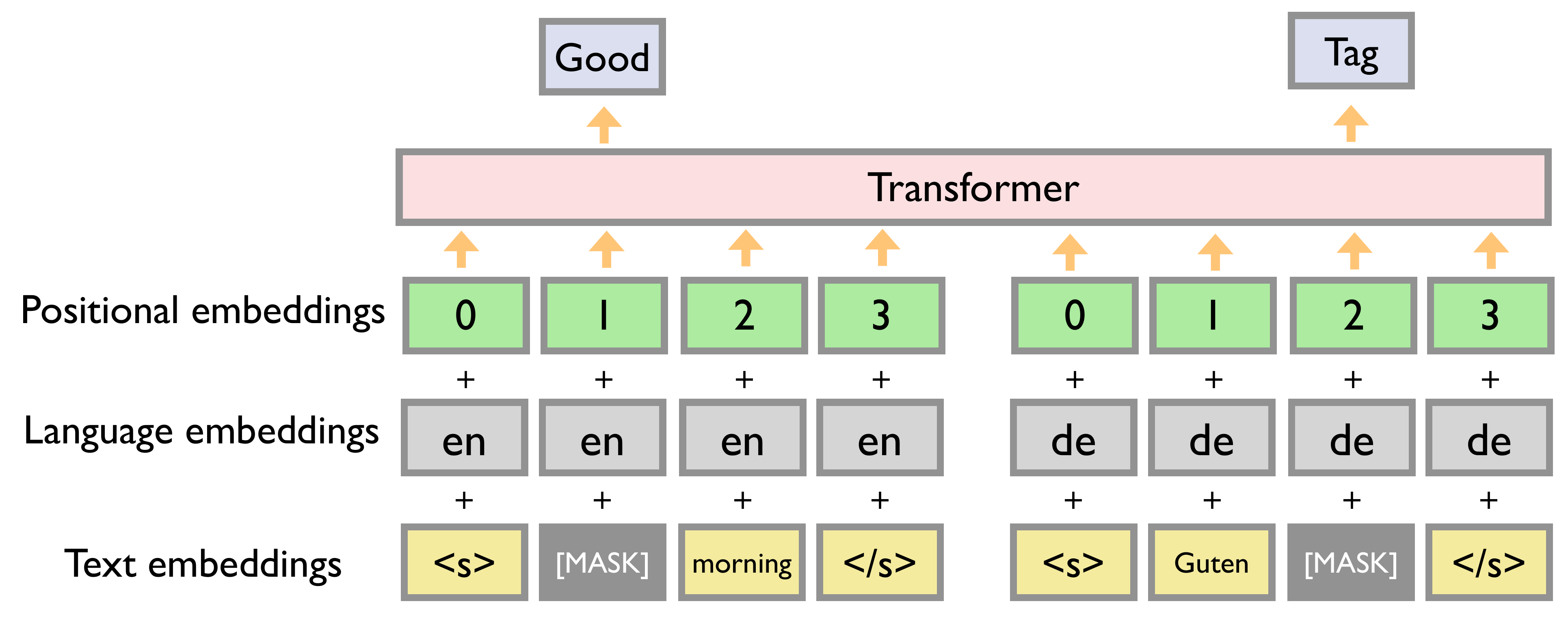}
\label{fig:tlm}
}
\end{center}
\end{minipage}
\begin{minipage}[b]{0.45 \linewidth}
\begin{center}
\subfigure[Masked Acoustic Model (MAM) for speech.]{
\includegraphics[width=6cm]{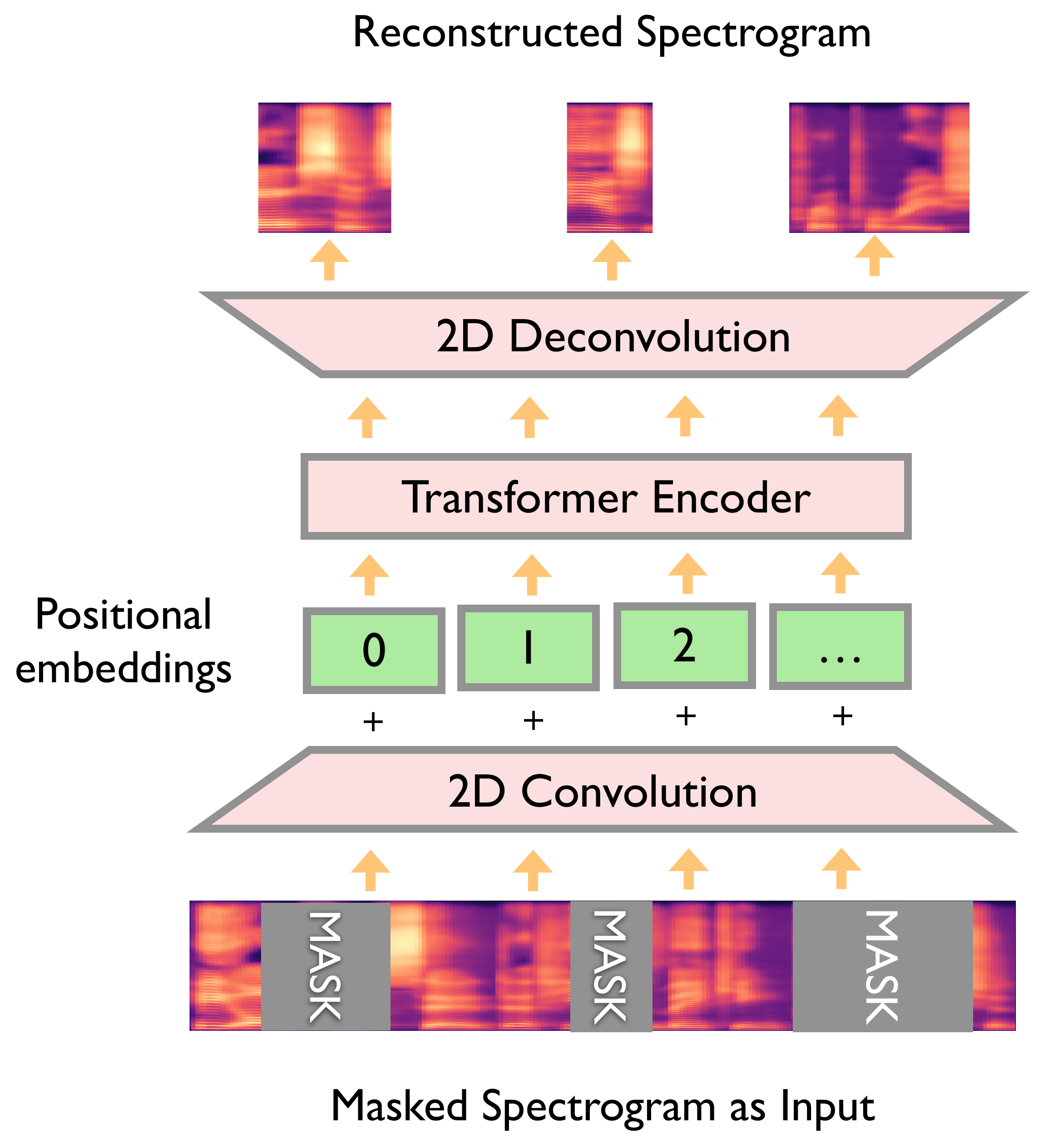}
\label{fig:mam}
}
\end{center}
\end{minipage}

\end{tabular}
\caption{Previous work for speech or text monomodal
representation learning.
}
\vspace{-0.5cm}
\label{fig:exist_work}
\end{figure*}

Inspired by recent cross-lingual language model pre-training
work \cite{lample2019cross} which shows the potential to
unify the representations of different languages into
one encoder,
we propose a Fused Acoustic and Text Masked Language Model 
(FAT-MLM).
This model jointly learns a unified representation 
for both acoustic and text input.
In this way, we extend the masked language model's input
from only acoustic or text data to multimodal corpora containing
both acoustic and text data,
such as speech recognition
and speech translation for the first time (Fig.~\ref{fig:intro}).

We further extend this Fused Acoustic and Text encoder
to a sequence-to-sequence framework and
present an end-to-end Speech Translation model 
(FAT-ST).
This enables the model to be trained from both speech and text machine translation data
into one single encoder-decoder model.
Meanwhile, this model can also learn from speech recognition
data using an extra FAT-MLM loss.
This resolves the limitation of existing single encoder and decoder
speech translation models, which can only learn from
scarce parallel speech translation data,
but neglects much larger scale speech recognition and
text machine translation data (Fig.~\ref{fig:intro}).

We make the following contributions:

\begin{itemize}
    \item We propose the Fused Acoustic and Text Masked 
    Language Model (FAT-MLM), which can learn a unified 
    acoustic and text representation.
    
    \item Based on FAT-MLM, we propose the
    Fused Acoustic and Text Speech Translation model (FAT-ST),
    which can do speech recognition and machine translation
    in a single encoder-decoder framework. 
    
    \item Spontaneous speech translation experiments on three 
    language pairs 
    show that by finetuning  FAT-MLM, 
    the accuracy of FAT-ST improves end-to-end speech
    translation model by $+4.65$ BLEU in average and achieves
     state-of-the-art.
    This is the first time that an end-to-end speech translation model achieves
    similar performance with the strong cascaded system
    in these three translation directions
    of this dataset, while still maintaining a smaller 
    model size and faster decoding time. 
    
    \item We show that FAT-MLM trained with additional speech recognition,
    machine translation, and monolingual text data can improve FAT-ST
    by $+1.25$ BLEU.
    FAT-ST can be further improved by using additional speech recognition
    and machine translation data.

\end{itemize}

\section{Previous Work}



\subsection{Masked Language Modeling}

\citet{radford2018improving},
\citet{howard2018universal} and \citet{BERT}
investigate language modeling for pretraining Transformer encoders. 
Unlike \citet{radford2018improving} using 
unidirectional language models
for pretraining, \citet{BERT} proposes BERT which enables 
deep bidirectional representation pretraining by a
masked language modeling (MLM) objective 
inspired by the Cloze task \cite{taylor1953cloze}
which randomly masks some of the tokens from the input,
with an objective to recover the masked
word based only on its context.
Their approaches lead to drastic improvements on several 
natural language understanding tasks including
text classification \cite{wang2018glue},and
question answering \cite{rajpurkar2016squad}.

\subsection{Translation Language Modeling}

\begin{figure*}[ht!]
\centering
\begin{tabular}{cc}
\\
\begin{minipage}[b]{0.35 \linewidth}
\begin{center}
\subfigure[Speech Reconstruction Module.]{
\includegraphics[width=4.5cm]{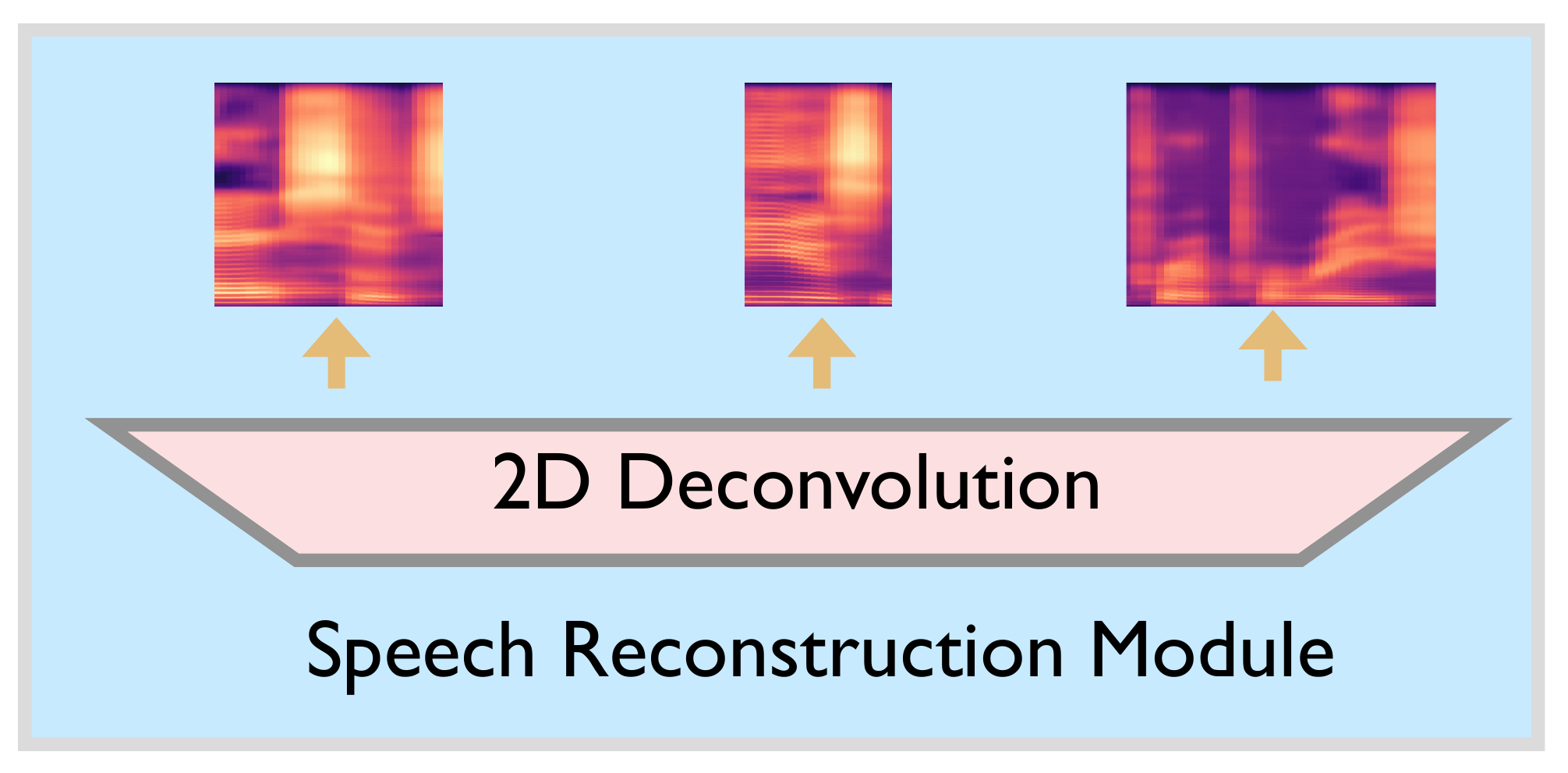}
\label{fig:fat_acoustic_recover}
}
\end{center}
\end{minipage}
\begin{minipage}[b]{.65 \linewidth}
\begin{center}
\subfigure[Monolingual FAT-MLM.]{
\includegraphics[width=7.5cm]{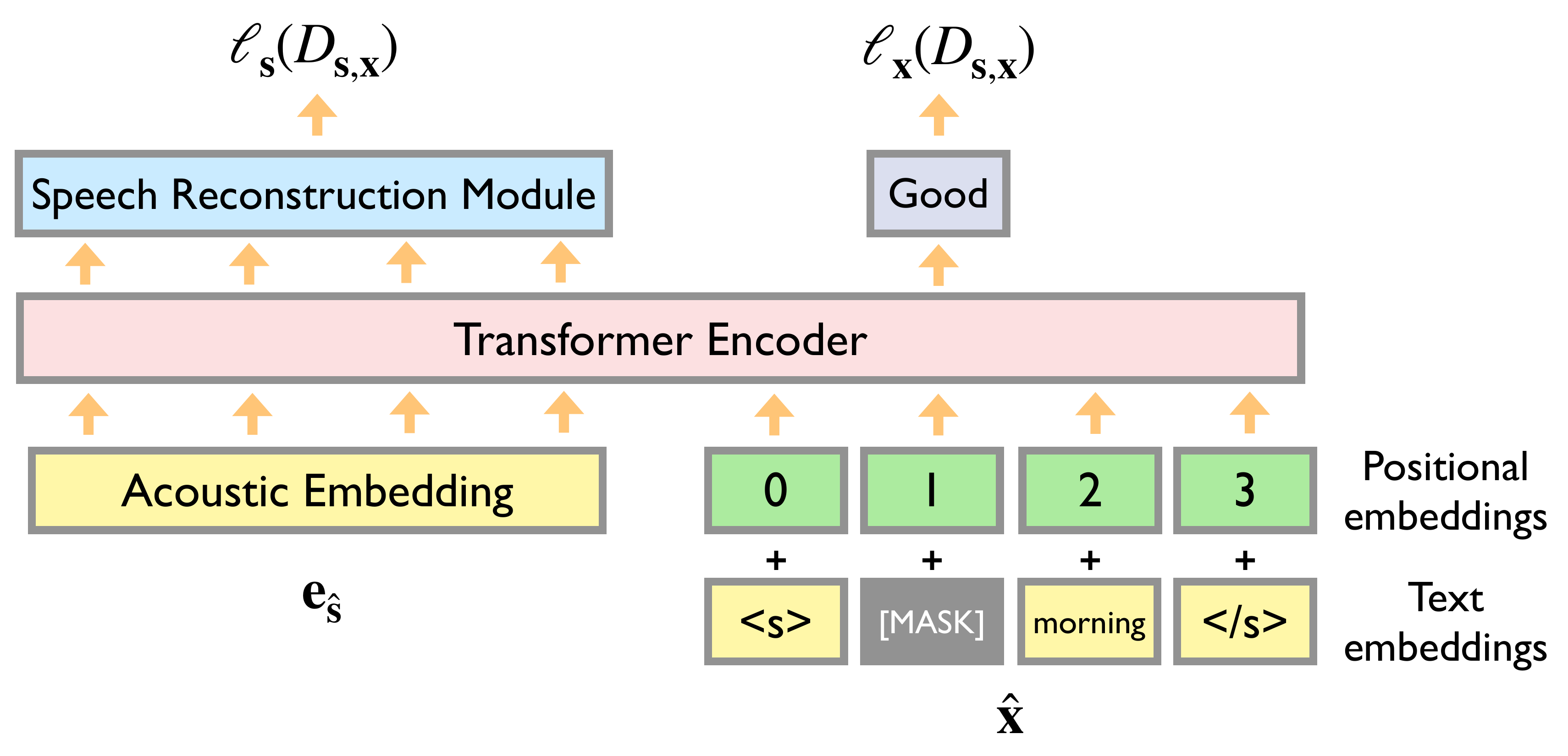}
\label{fig:fat_mlm_st}
}
\end{center}
\end{minipage}

\\

\begin{minipage}[b]{0.35 \linewidth}
\begin{center}
\captionsetup[subfigure]{width=\linewidth}
\subfigure[Acoustic Embedding Module.]{
\includegraphics[width=4.5cm]{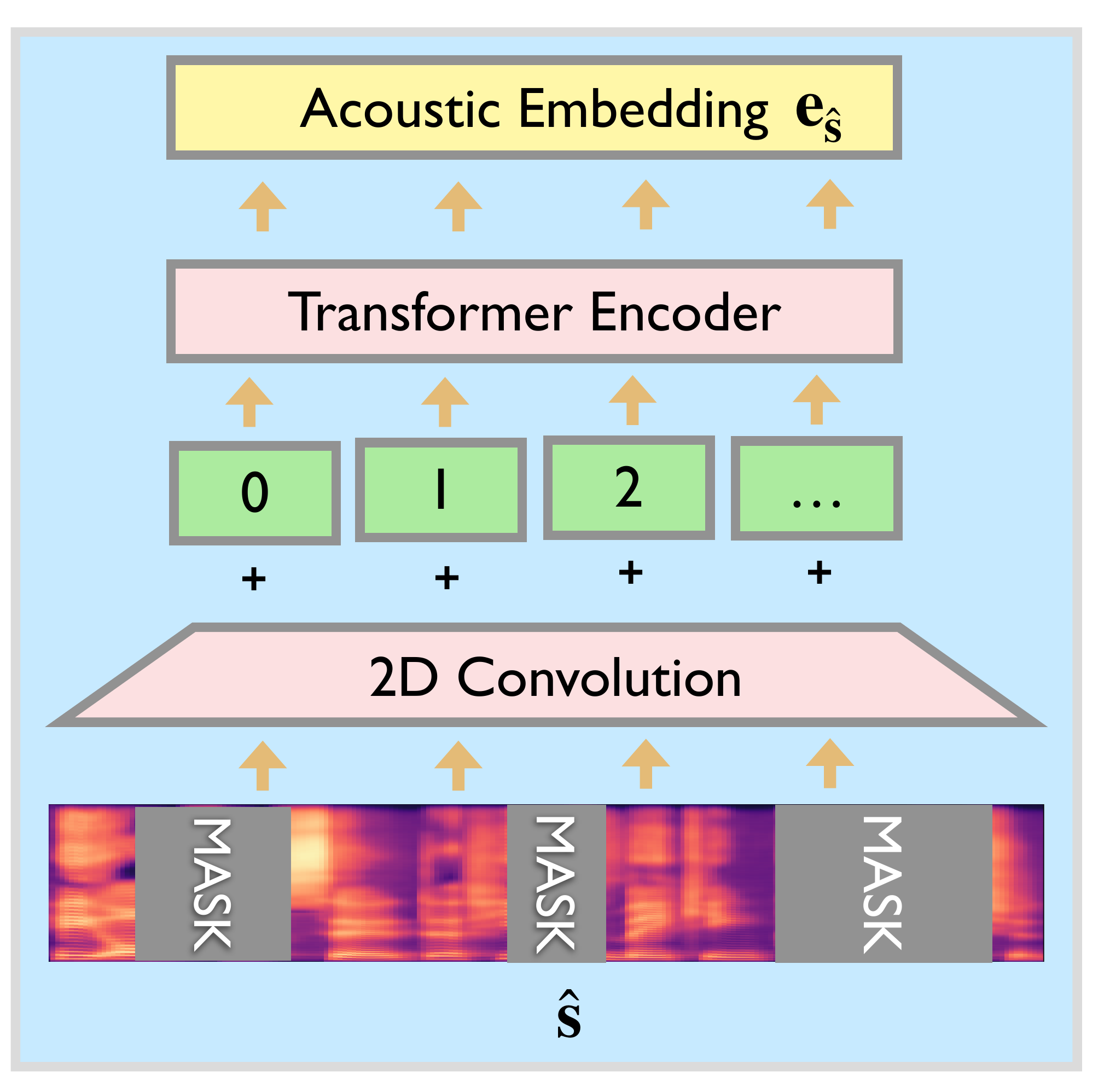}
\label{fig:fat_acoustic_embed}
}
\end{center}
\end{minipage}
\begin{minipage}[b]{.65 \linewidth}
\begin{center}
\subfigure[Translation FAT-MLM]{
\includegraphics[width=11cm]{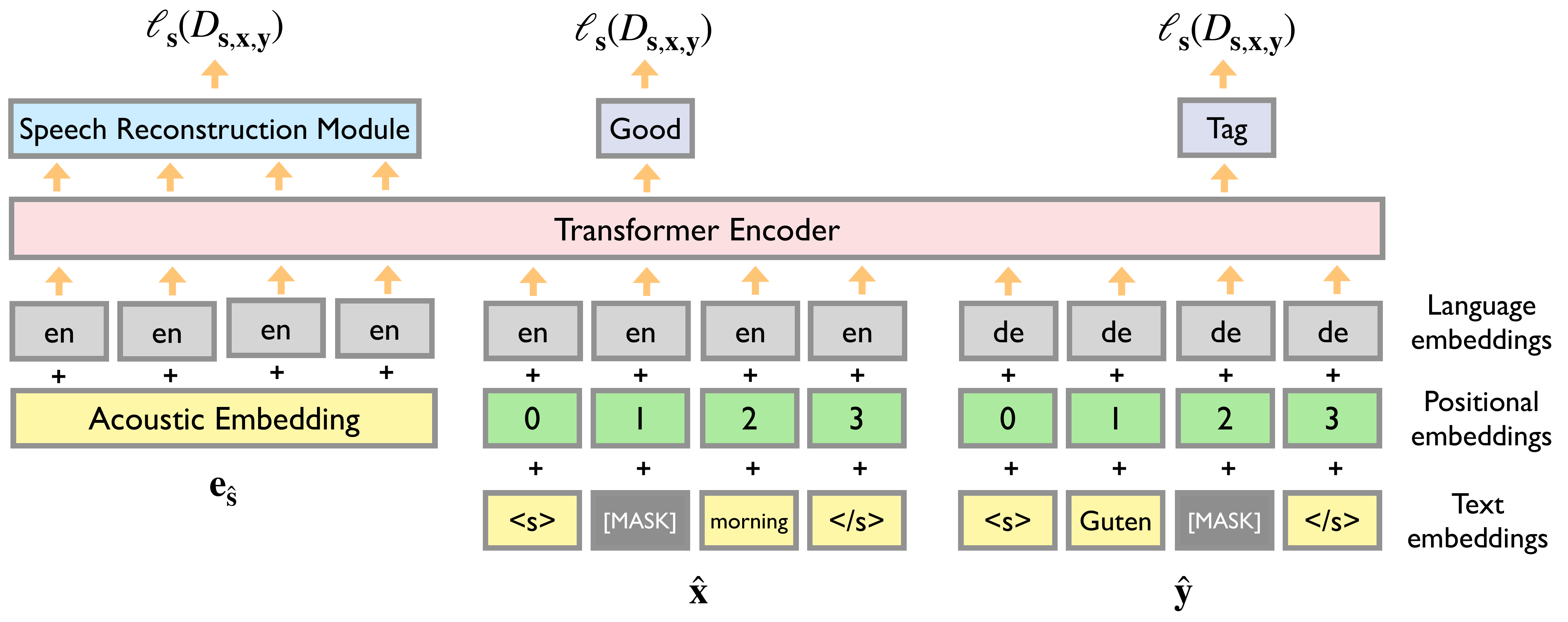}
\label{fig:fat_mlm_st}
}
\end{center}
\end{minipage}
\end{tabular}
\caption{Fused Acoustic and Text-Masked Language Model (FAT-MLM).}
\label{fig:fat_mlm}
\end{figure*}

\citet{lample2019cross} extend MLM to cross-lingual pretraining
by proposing two methods:
one unsupervised that only relies on monolingual data,
and one supervised that leverages parallel data with a new 
cross-lingual language model objective
which is called Translation Language Model (TLM). 
As shown in Fig.~\ref{fig:tlm}, 
TLM encodes both source and target sentences from a parallel data
after masking several tokens with \texttt{[MASK]}, 
and then learn to recover the masked tokens.
Experiments show that TLM achieves state-of-the-art results on 
cross-lingual classification, unsupervised and supervised machine translation.

\subsection{Masked Acoustic  Model}

Recently, \citet{chen2020mam}
propose to learn a speech encoder in a self-supervised fashion 
on the speech side, which can utilize speech data without transcription.
This technique termed Masked Acoustic Modeling (MAM), can also perform pretraining
on any acoustic signals (including non-speech ones) without annotation.
Fig.~\ref{fig:mam} demonstrate the architecture of MAM.
Similar with MLM, MAM replaces a span of speech spectrogram with mask tokens
\texttt{[MASK]}. 
After a 2D Convolution layer and a Transformer Encoder, MAM learns to recover
the masked spectrogram via a 2D De-convolution layer during training.
\citet{chen2020mam} shows that MAM can improve
end-to-end speech translation as either an additional loss or
a pretraining model. 
Parallel to MAM, \citet{baevski2020wav2vec}
proposes the wav2vec 2.0 pretraining model, which masks the speech input 
in the latent space and pretrains the model via a contrastive task defined
over a quantization of the latent representations.

\section{Fused Acoustic and Text Masked Language Model (FAT-MLM)}


Although existing pretraining models show a strong representation learning ability 
and significantly improve upon many down-streaming tasks,
they all can only learn the representation for 
either text or speech.
However, a unified speech and text multi-modal representation is useful 
for many end-to-end spoken language processing tasks.

To address this problem, we propose the Fused Acoustic and Text Masked
Language Model (FAT-MLM), a multimodal pretraining model
which encodes acoustic, text into a unified representation.
The idea is similar with \citet{lample2019cross} who
propose to learn a unified representation
of different languages.
They first propose a method relying on
the shared sub-word vocabulary to align
different languages' representation.
However this is unapplicable in our case
because of the modality difference.
Thus we propose a method similar to
their second approach
TLM which uses parallel speech recognition
data.
In the following sections, we first introduce 
the monolingual FAT-MLM and then show how
to extend it to translation scenario.

\subsection{Monolingual FAT-MLM}

The monolingual FAT-MLM takes speech and
transcription tuples as input, denotes as $D_{\vecs, \vecx} = \{(\vecs, \vecx)\}$, 
where $\vecs=(s_1, ... , s_{|s|})$ is a sequence of 
acoustic features 
 $s_i \in \mathbb{R}^{d_s}$ 
which 
can be the spectrogram 
or mel-spectrogram of the 
speech audio,
and each $s_i$ represents the frame-level speech 
feature, 
and $\vecx=(x_1, ... , x_{|\vecx|})$ is the sequence of corresponding transcription.

As shown in Fig.~\ref{fig:fat_acoustic_embed}, 
we first
randomly mask several spans of $\vecs$
by a random masking function over the 
input $\vecs$:
\begin{equation}
\hat{\vecs} \sim  \text{Mask}_{\text{span}}(\vecs, \lambda)
\label{eq:randx}
\end{equation}
where $\text{Mask}_{\text{span}}(\cdot)$ 
replaces several random spans of $\vecs$
by probability of $\lambda$ (30\% in our work)
with a random initialized vector $\epsilon_{\vecs} \in \mathbb{R}^{d_\vecs}$.
Then we encode $\hat{\vecs}$ with Convolutions and a  Transformer encoder
for acoustic embeddings $\vece_{\hat{\vecs}}$.
Similarly, we randomly mask tokens in $\vecx$
by a random masking function over the 
input $\vecs, \vecx$:
\begin{equation}
\vspace{-5pt}
\hat{\vecx} \sim  \text{Mask}_{\text{token}}(\vecx, \lambda)
 \vspace{-1pt}
\label{eq:randx}
\end{equation}
where $\text{Mask}_{\text{token}}(\cdot)$ 
replaces several tokens of $\vecx$
by probability of $\lambda$
with a random initialized vector $\epsilon_{\text{token}} \in \mathbb{R}^{d_{\vecx}}$.
Then we concatenate acoustic embeddings 
and source text embeddings 
$[\hat{\vece}_{\vecs}; \hat{\vecx}]$,
and 
obtain the
latent representation
$f([e_{\hat{\vecs}}; \hat{\vecx}])$
using another
Transformer encoder, denoted as $f$.
Same with \citet{lample2019cross}, we reset the
positional embeddings for different types of input.

The training objective of monolingual FAT-MLM
includes a speech reconstruction loss $\ell_{\vecs}(\dsx)$
and a text reconstruction loss $\ell_{\vecx}(\dsx)$.
For speech input $\vecs$,
we have the following training objective to reconstruct the 
original speech signal with the surrounding context 
information\footnote{Similar with previous work 
using masked
language model objective,
this loss only takes the masked input into consideration.}:
\begin{equation}
\ell_{\vecs}(D_{\vecs, \vecx}) = \textstyle\sum_{(\vecs, \vecx) \in D_{\vecs, \vecx}} || \vecs - g (f ([e_{\hat{\vecs}}; \hat{\vecx}]) ||_2^2
\label{eq:reconstruct}
\end{equation}
where $g$ is a reconstruction function 
(we use 2D deconvolution in this work)
which tries to recover the original
signal from encoded representation $f([e_{\hat{\vecs}}; \hat{\vecx}])$.
We use
mean squared error for measuring the difference
between $s$ and the reconstructed spectrogram.
For transcription input $\vecx$,
following \citet{BERT} we use  cross entropy loss
, denoted as 
\begin{equation}
\ell_{\vecx}(\dsx) = - \textstyle\sum_{(\vecs,\vecx)\in \dsx} \log p(\vecx \mid [\vece_{\hat{\vecs}}; \hat{\vecx}]) 
\label{eq:train}
\end{equation}
to reconstruct the masked token. 
The final loss for monolingual FAT-MLM is:
\begin{equation}
\ell_{\text{FAT-MLM}}(D_{\vecs, \vecx}) = \ell_{\vecs}(D_{\vecs, \vecx}) + \ell_{\vecx}(D_{\vecs, \vecx})
\vspace{-0.2cm}
\label{eq:fatmlm-loss}
\end{equation}

\subsection{Translation FAT-MLM}

To support multimodal
crosslingual tasks such as speech translation,
We propose Translation FAT-MLM 
which extends Monolingual FAT-MLM by 
using additional target language translation of 
the source language transcription as input.
Formally Translation FAT-MLM takes 
$D_{\vecs, \vecx, \vecy} = \{(\vecs, \vecx, \vecy)\}$
as input, 
where $\vecy= [y_1, ... , y_{|y|}]$ denotes the sequence of 
target language translation. 
This kind of triplet input is very common in speech translation corpus.

As shown in Fig.~\ref{fig:fat_mlm_st}, 
we incorporate source language embedding $e_{\text{src}}$
and target language embedding $e_{\text{tgt}}$
for different languages
to show the language difference.
Similar to Monolingual FAT-MLM, 
Translation FAT-MLM randomly masks the translation
input $\hat{\vecy} \sim  \text{Mask}_{\text{token}}(\vecy, \lambda)$
and concatenate it with another two embeddings:
$$\vech_{\vecs, \vecx, \vecy} = [\vece_{\hat{\vecs}}+\vece_{\text{src}}; \hat{\vecx}+\vece_{\text{src}}; \hat{\vecy}+\vece_{\text{tgt}}]$$
Then we reconstruct masked input from concatenated embeddings
$\vech_{\vecs, \vecx, \vecy}$ via a Transformer encoder.
The reconstruction loss 
for different masked input is:
\begin{equation}
\notag
\begin{split}
\ell_{\vecs}(\dsxy) & = \textstyle\sum_{(\vecs, \vecx, \vecy) \in \dsxy} || \vecs - g (f (\vech_{\vecs, \vecx, \vecy}) ||_2^2 \\
\ell_{\vecx}(\dsxy) & = - \textstyle\sum_{(\vecs,\vecx, \vecy)\in \dsxy} \log p(\vecx \mid \vech_{\vecs, \vecx, \vecy}) \\
\ell_{\vecy}(\dsxy) & = - \textstyle\sum_{(\vecs,\vecx, \vecy)\in \dsxy} \log p(\vecy \mid \vech_{\vecs, \vecx, \vecy} )
\label{eq:train}
\end{split}
\end{equation}
We sum these loss functions for the final loss function of
Translation FAT-MLM:
\begin{equation}
\notag
\ell_{\text{FAT-MLM}}(D_{\vecs, \vecx, \vecy}) = \ell_{\vecs}(D_{\vecs, \vecx, \vecy}) +
\ell_{\vecx}(D_{\vecs, \vecx, \vecy}) +
\ell_{\vecy}(D_{\vecs, \vecx, \vecy})
\label{eq:fatmlm-loss}
\end{equation}

To fully utilize the corpora for different tasks,
FAT-MLM can take any combination of speech, transcription,
translation triplets $D_{2^{\{\vecs, \vecx, \vecy\}}}$
as input.\footnote{$2^{\{\vecs, \vecx, \vecy\}}$ is the power set
of $\{\vecs, \vecx, \vecy\}$ triplets.}
Specifically, these combinations include
speech only data $\{ \vecs \}$, 
monolingual text data, $\{ \vecx \}$ or $\{ \vecy \}$,
speech and transcription tuple  
$\{ (\vecs, \vecx) \}$
for speech recognition, 
transcription and translation tuple 
$\{ (\vecx, \vecy) \}$ 
for machine translation,
speech and translation tuple 
$\{ (\vecs, \vecy) \}$ 
for direct speech translation
and speech transcription translation triplets
$\{ (\vecs, \vecx, \vecy) \}$.
For different combinations of input, FAT-MLM
encodes the full concatenation of their embeddings
and
recover the masked portion.
The loss function is:
\begin{equation}
\ell_{\text{FAT-MLM}}(D_{2^{\{\vecs, \vecx, \vecy\}}}) = \ell_{\vecs}(D_{\vecs\star}) +
\ell_{\vecx}(D_{\vecx\star}) +
\ell_{\vecy}(D_{\vecy\star})
\label{eq:fatmlm-loss_all}
\end{equation}
where $D_{\vecs\star}$, $D_{\vecx\star}$, $D_{\vecy\star}$
means any input including speech, source language text
and target language text respectively.
Note that in this framework, 
we can denote MLM as $\ell_{\vecx}(\dx)$,
TLM as $\ell_{\vecx, \vecy}(\dxy)$,
MAM as $\ell_{\vecs}(\vecs)$.

\subsection{Attention Visualization}

\begin{figure}[tb!]
\centering
\includegraphics[width=4cm]{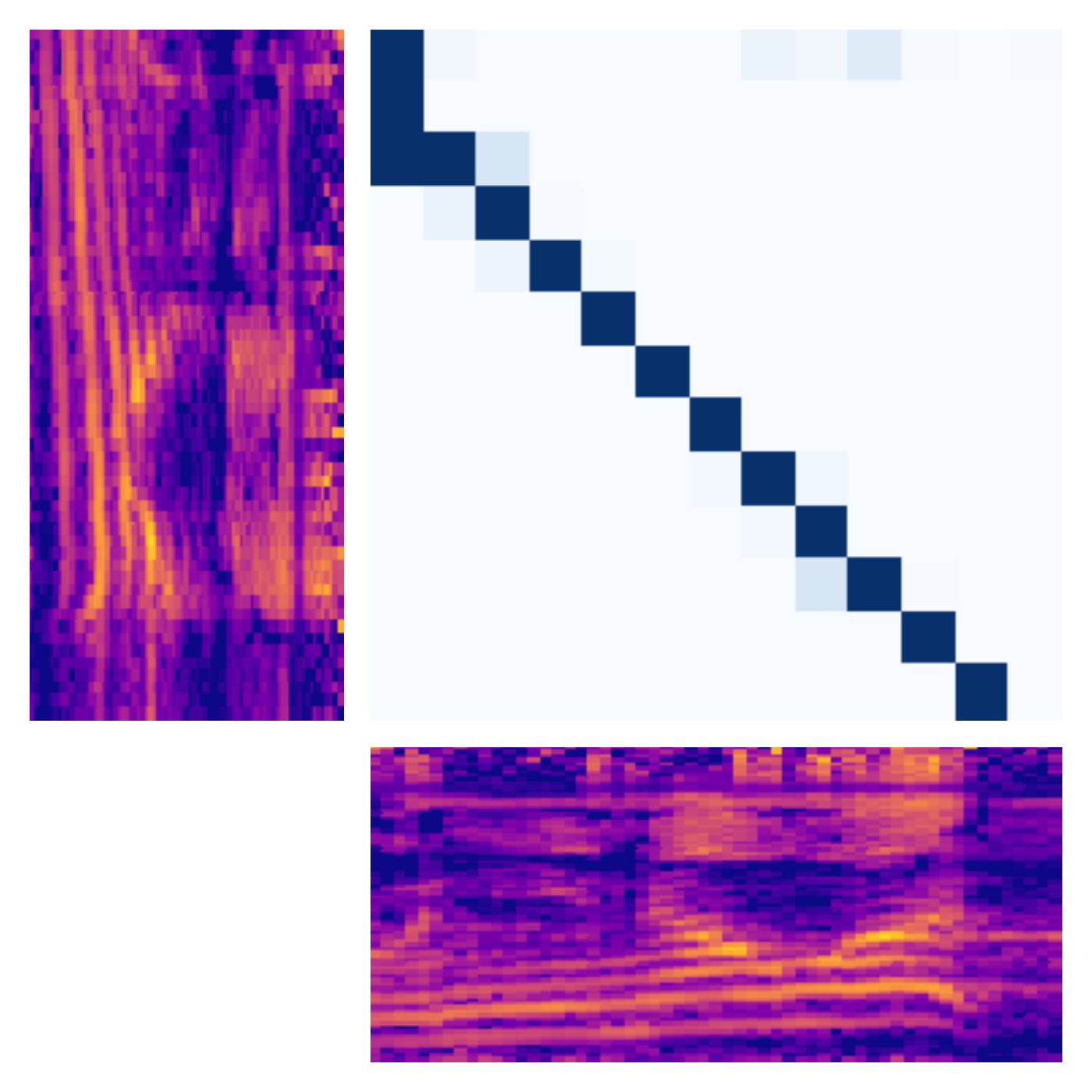}
\vspace{-.2cm}
\caption{
One speech self-attention head's output at the first transformer layer in acoustic embedding module
and its corresponding spectrogram.
This is a Translation FAT-MLM model trained 
with Must-C
En$\to$De dataset.
}
\label{fig:attn_speech}
\end{figure}


\begin{figure}[ht!]
\centering
\begin{tabular}{c}
\\
\begin{minipage}[t]{.9 \linewidth}
\begin{center}
\subfigure[
This self-attention head shows  
bilingual alignment
between
``\texttt{and}'‘ and
``\texttt{Und}'',
``\texttt{you}'‘ and
``\texttt{Sie}'',
``\texttt{what}''  and
``\texttt{?}'' in
transcription and translation respectively.
]{
    \makebox[\linewidth][c]{
\includegraphics[height=5cm]{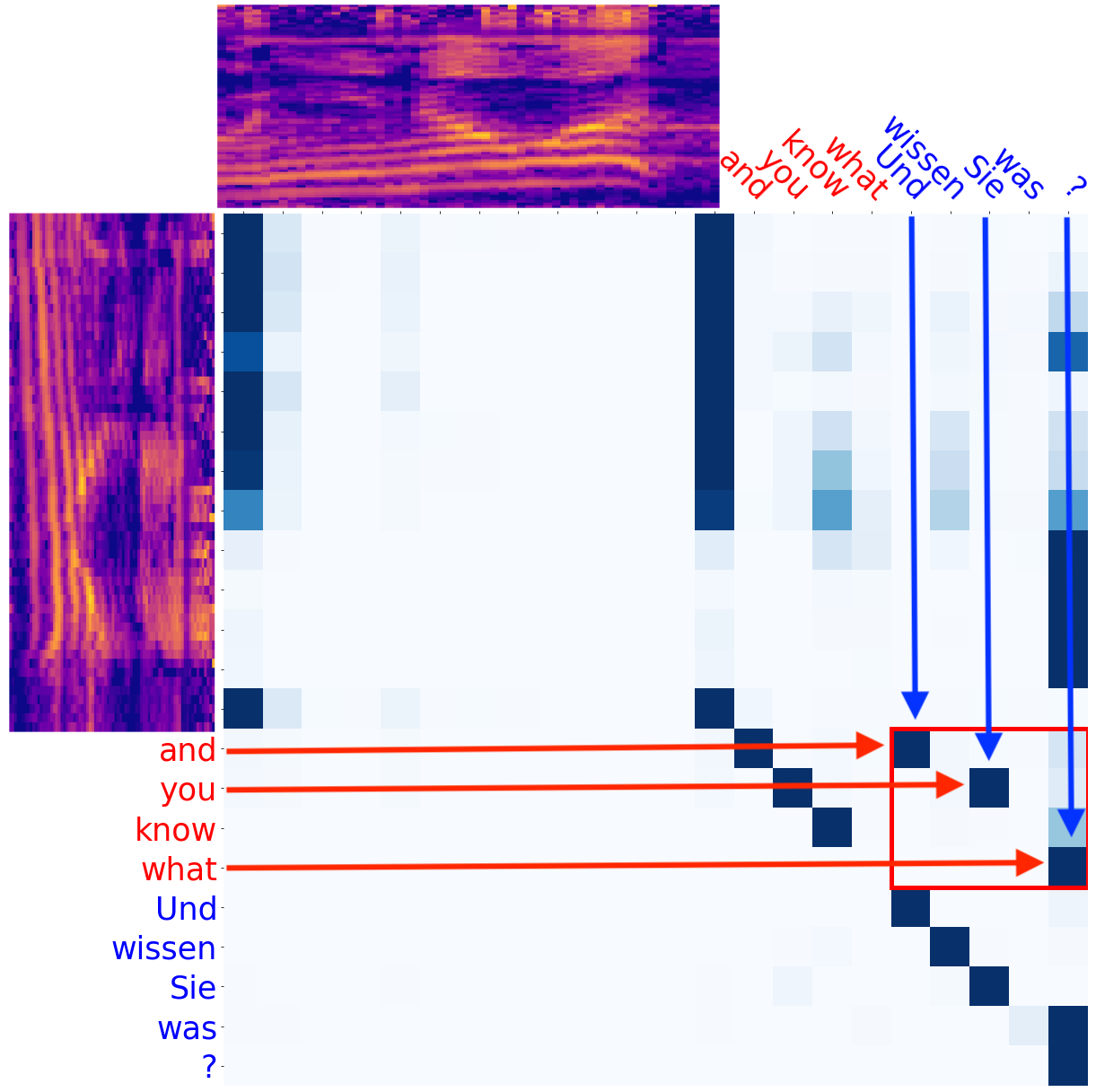}
\label{fig:attn_translation}
    }
}
\end{center}
\end{minipage}
\\

\begin{minipage}[t]{.9 \linewidth}
\begin{center}
\subfigure[
Left side spectrogram shows gold 
speech-transcription alignment.
This self-attention head shows monotonic 
crossmodal attention 
in red box.
Meanwhile, the speech-to-translation
attention (in blue box) clearly
show the alignment between
``\texttt{you}'‘ and
``\texttt{Sie}'',
``\texttt{know}''  and
``\texttt{wissen}'' in
speech and translation respectively.
Note that in this speech, the
pronounciation of ``\texttt{and}''
is very weak.
]{
\makebox[\linewidth][c]{
\includegraphics[height=6cm]{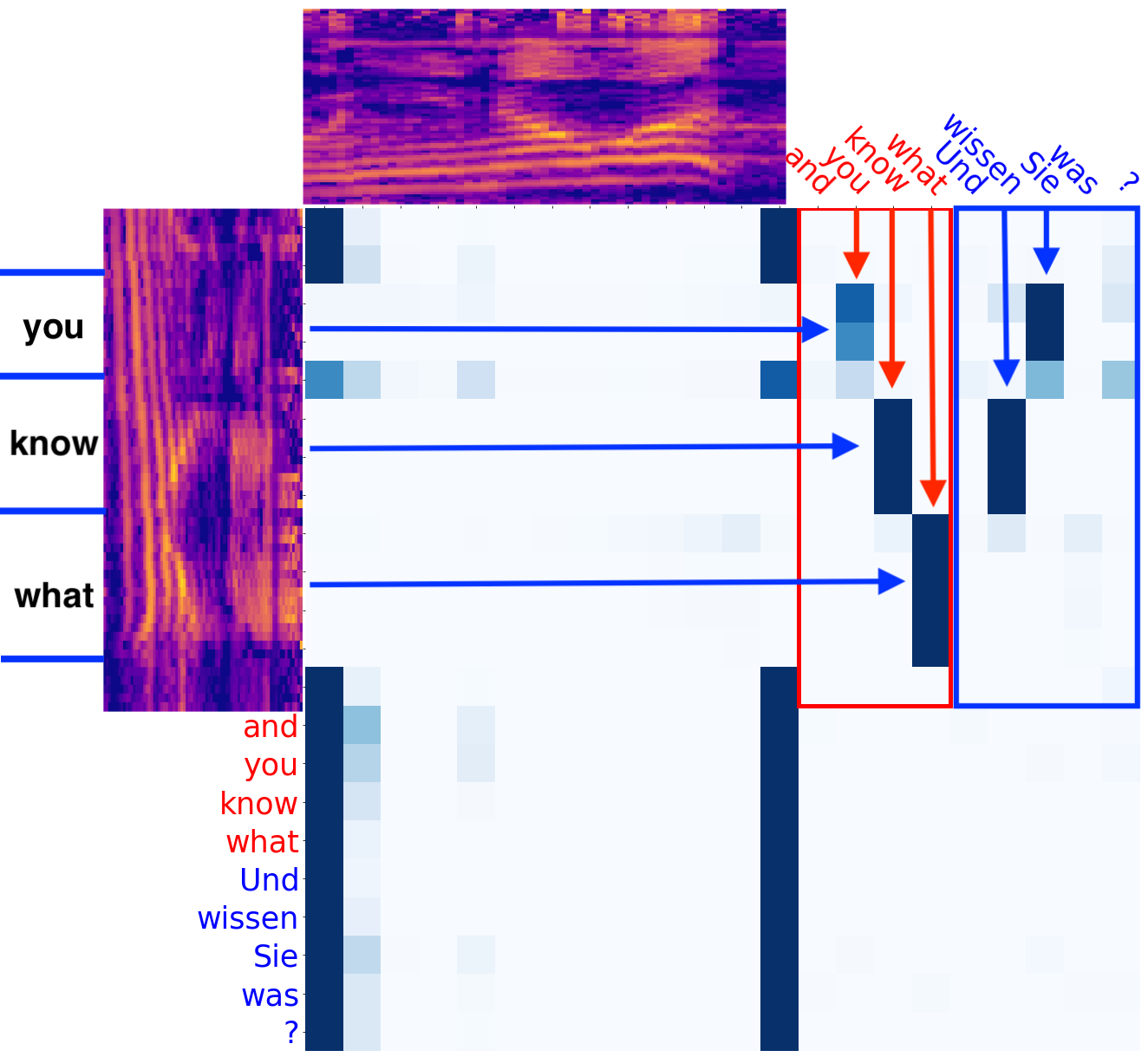}
\label{fig:attn_asr}
    }
}
\end{center}
\end{minipage}

\end{tabular}
\caption{
Two self-attention heads' output 
at the first layer of acoustic and text shared 
transformer
from a Translation FAT-MLM model trained with Must-C
En$\to$De dataset,
annotated with corresponding spectrogram, 
transcription (red) and translation (blue).
}
\label{fig:attn}
\end{figure}

\begin{figure}[t!]
\centering
\includegraphics[width=.8\linewidth]{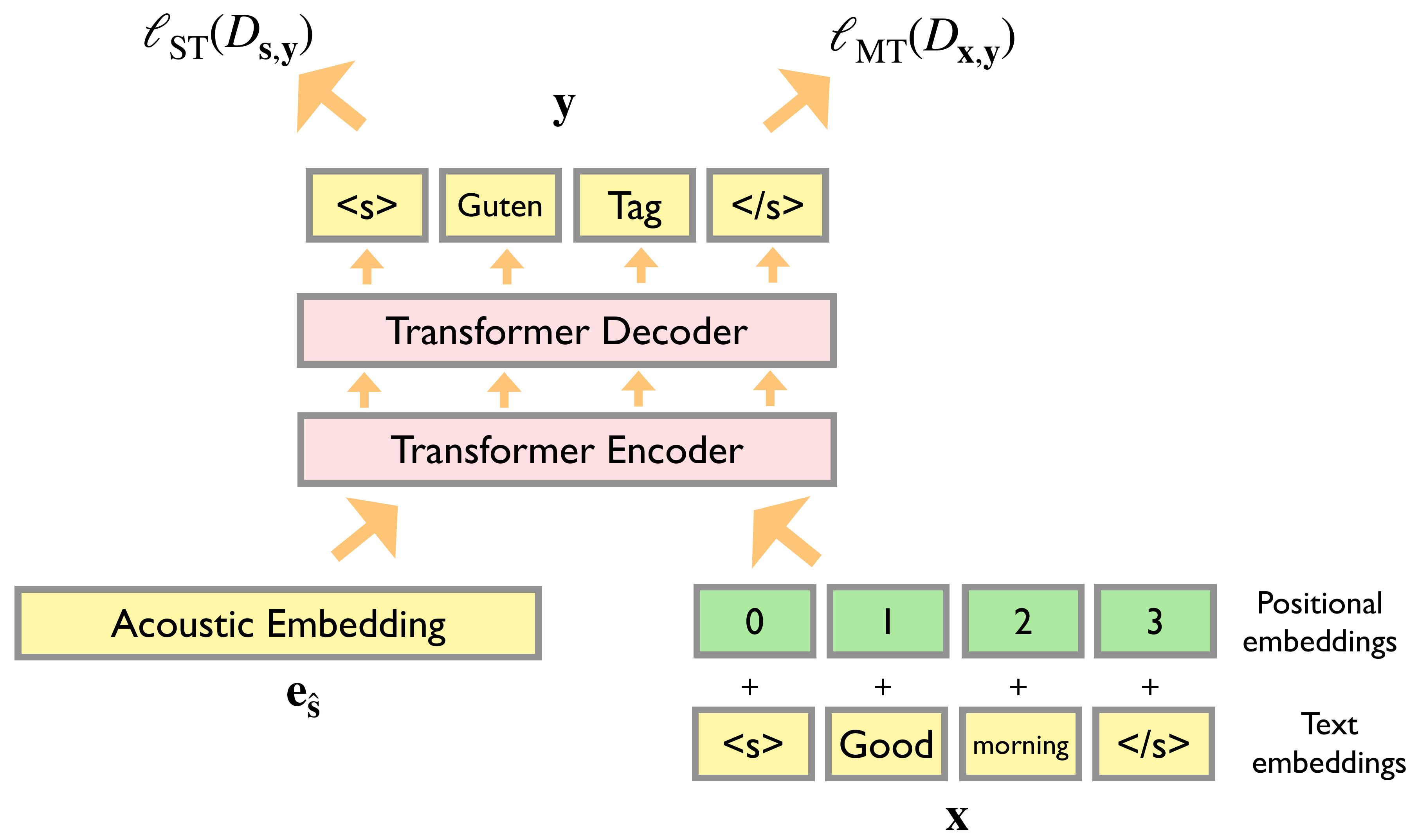}
\caption{Fused Acoustic and Text-Speech Translation (FAT-ST).}
\label{fig:fat_st}
\end{figure}

To demonstrate FAT-MLM's ability to unify the representation of
different modality and language, 
we show the self-attention layers of a
translation FAT-MLM
in Fig.~\ref{fig:attn_speech} and ~\ref{fig:attn}.
The clear monotonic attention
in Fig.~\ref{fig:attn_speech} shows
that our proposed method can learn good 
representation for speech \cite{chen2020mam}.
Fig.~\ref{fig:attn_translation} shows that FAT-MLM can learn 
a good crosslingual alignment between two languages, 
such as \texttt{and} to \texttt{Und} and \texttt{you} to \texttt{Sie}.
Fig.~\ref{fig:attn_asr} shows that 
FAT-MLM is able to learn
a clear monotonic speech-to-text crossmodal attention
like many speech recognition models.

\section{Fused Acoustic and Text Speech Translation (FAT-ST)}


In this section, we present how to adapt FAT-MLM to speech
translation and enable speech translation models to learn from
speech recognition and text machine translation.

\subsection{From Text Translation to Speech Translation}

Regardless of the particular design of different seq-to-seq models,
the text machine translation encoder always takes the input 
sequence $\vecx = (x_1,...,x_n)$ 
where each $x_i \in \mathbb{R}^{d_x}$ is a word embedding of $d_x$ dimensions,
and produces a new sequence of hidden states $\vech =f(\vecx) = (h_1,...,h_n)$. 
On the other hand, a decoder 
predicts 
the next output word $y_t$
given the source sequence (actually its representation \vech) 
and previously generated words, denoted $\vecy_{<t}=(y_1,...,y_{t-1})$.
The decoder 
stops when it emits \eos,
and the final hypothesis $\vecy = (y_1,...,\eos)$ 
has probability 
\begin{equation}
p(\vecy \mid \vecx)_{\text{MT}} = \textstyle\prod_{t=1}^{|\vecy|}  p(y_t \mid \vecx,\, \vecy_{<t})
\label{eq:gensentscore}
\end{equation}
At training time, we maximize the conditional probability of each ground-truth target sentence $\vecy^\star$ given input $\vecx$
over the whole training data $\dxy$, or equivalently minimizing the following loss:
\begin{equation}
\ell_{\text{MT}}(D_{\vecx, \vecy}) = - \textstyle\sum_{(\vecx,\vecy)\in D_{\vecx, \vecy}} \log p(\vecy \mid \vecx) 
\label{eq:train}
\end{equation}

Different from text machine translation, speech translation
takes speech features $\vecs = (s_1, ..., s_{|\vecs|})$ as input.
Same as the speech input portion of FAT-MLM, these speech features
are converted from the speech signals (e.g. spectrogram).
Formally, the decoding and training of speech translation models
can be defined as follows:
\begin{equation}
p(\vecy \mid \vecs)_{\text{ST}} = \textstyle\prod_{t=1}^{|\vecy|}  p(y_t \mid \vecs,\, \vecy_{<t})
\label{eq:gensentscore}
\vspace{-0.2cm}
\end{equation}
\begin{equation}
\ell_{\text{ST}}(\dsy) = - \textstyle\sum_{(\vecs,\vecy)\in \dsy} \log p(\vecy \mid \vecs) 
\label{eq:train}
\end{equation}

\subsection{FAT-ST}

To boost the performance of end-to-end speech translation,
we propose to enable speech translation to encode both acoustic
and text features as
input by simply adapting the architecture of monolingual FAT-MLM to
a Fused Acoustic and Text Speech Translation model (FAT-ST).

FAT-ST's encoder shares
identical architecture with monolingual FAT-MLM.
In this way, we can simply encode either acoustic or text features
by this encoder
and the FAT-ST model can be optimized by speech translation loss
$\ell_{\text{ST}}$, machine translation loss $\ell_{\text{MT}}$
and FAT-MLM loss $\ell_{\text{FAT-MLM}}$.
For a speech translation dataset $\dsxy$,
we decouple the triplets into three part
$\dsy$ for $\ell_{\text{ST}}$, 
$\dsx$ for $\ell_{\text{FAT-MLM}}$
and $\dxy$ for $\ell_{\text{MT}}$.
The loss function of FAT-ST is:
\begin{equation}
\begin{split}
\notag
\ell_{\text{FAT-ST}}(\dsy \cup \dsx \cup \dxy) = \ell_{\text{ST}}(\dsy) 
+  \ell_{\text{MT}}(\dxy) \\
+  \ell_{\text{FAT-MLM}}(\dsx) 
\label{eq:fatst_loss}
\end{split}
\end{equation}
Please note that the speech recognition and machine translation data
can either be included in speech translation
data 
or additional datasets.
Meanwhile, in practice, we find that CTC loss \cite{graves2006connectionist}
is useful to improve the translation quality
so that we include it in all the experiments.

\subsection{Finetuning FAT-ST from Translation FAT-MLM}


Similar to \citet{lample2019cross} we can further improve FAT-ST by 
finetuning from FAT-MLM.
Since the FAT-ST decoder predicts text only,
we initialize it from the
acoustic and text shared Transformer encoder.
Although Transformer decoder is unidirectional
which is different from bidirectional FAT-MLM,
it can still benefit from FAT-MLM in our experiments,
This is also observed by \citet{lample2019cross}
and \citet{BERT}.

\section{Experiments}


We conducted speech translation experiments
in 
3 directions: English to German (En$\to$De), 
English to Spanish (En$\to$Es), and English to Dutch (En$\to$Nl) to show the translation quality of
baselines and our proposed
methods.

\subsection{Dataset}


\begin{table}[ht!]
	\centering
	\begin{subtable}[Bilingual Dataset]{
\resizebox{1.0\columnwidth}{!}{
		\begin{tabular}{l|l|cc|cc|cc}
\toprule
\multirow{2}{*}{Type} & \multirow{2}{*}{Name} & \multicolumn{2}{c|}{En $\to$ De}& \multicolumn{2}{c|}{En $\to$ Es}& \multicolumn{2}{c}{En $\to$ Nl} \\ 
    &      & Hours    & \#Sent           & Hours    & \#Sent       & Hours    & \#Sent    \\ \hline
$D_{\vecs,\vecx,\vecy}$ & Must-C ST & 408  & 226K  & 504 & 262K  & 442 & 245K        \\ 
$D_{\vecx,\vecy}$   &  Europarl MT & -  & 1.9M & -    & 2.0M  & -    & 2.0M \\
\midrule
\end{tabular}}
}
    \label{table:bi_data}
    \end{subtable}
	\begin{subtable}[Monolingual Dataset]{
		
\resizebox{1.0\columnwidth}{!}{
		\begin{tabular}{l|l|cc|c|c|c}
\toprule
 \multirow{2}{*}{Type} & \multirow{2}{*}{Name}  & \multicolumn{2}{c|}{En} & \multicolumn{1}{c|}{De} & \multicolumn{1}{c|}{Es} & \multicolumn{1}{c}{Nl} \\ 
    &                    & Hours   & \#Sent   &  \#Sent   &  \#Sent   &  \#Sent   \\ \hline
$D_{\vecs,\vecx}$ & Librispeech ASR     & 960     & 281K       &  -             &  -             &  -             \\ 
$D_{\vecs}$      & Libri-light  Speech     & 3,748    & 579K       &  -             &  -             &  -             \\ 
$D_{\vecx}/ D_{\vecy}$ & Europarl / Wiki Text & -   & 2.3M &  2.1M &  2.0M    &  2.3M        \\ 
\midrule
\end{tabular}
}
}\textbf{}\label{table:mono_data}
	\end{subtable}
	\caption{Statistics of all datasets used in our experiments. 
	Note that we use Europarl for En, De, Es monolingual text and 
	Wiki Text for Nl because there is no monolingual Nl portion in Europarl.
	\#Sent means the number of sentences.}\label{table:data}
\end{table}


We use 5 corpora with different modalities and languages:
speech translation data $\dsxy$ Must-C~\cite{mustc},
speech recognition data $\dsx$ Librispeech \cite{panayotov2015librispeech},
machine translation and monolingual text data 
$\dxy, \dx, \dy$ Europarl V7
\cite{koehn2005europarl},
speech only data $\ds$ Libri-Light (medium version)~\cite{librilight}
and monolingual text data Wiki Text (only for Nl).
The statistical results of the dataset are shown in Table.~\ref{table:data}.
We evaluate our models on Must-C dev and test set.
Note that Must-C is collected based on spontaneous
speeches (TED) which are very different from other
audiobook speech dataset used in our experiments.
Spontaneous speeches are
much harder for
speech translation than audiobook
dataset such as Libri-trans \cite{kocabiyikoglu2018augmenting}.
That is one of the reasons why
the translation accuracy of end-to-end
speech translation is much worse than
cascaded systems on Must-C than other
speech translation corpus.

\subsection{Training Detail}


\begin{table*}[ht]
\centering
\resizebox{1.5\columnwidth}{!}{
\begin{tabular}{@{}clcccccc@{}}
\toprule
Pretrain Method & Models & \multicolumn{1}{c}{En$\to$De} & \multicolumn{1}{c}{En$\to$Es} & \multicolumn{1}{c}{En$\to$Nl} & Avg. & Model Size \\ \midrule
\multirow{11}{*}{
\begin{tabular}[c]{@{}c@{}}No\\ Pretraining \end{tabular} 
}
& 
 ST                   & 19.64  & 23.68 & 23.01 & 22.11    &    31.25M\\
 & ST + ASR         & 21.70  & 26.83 & 25.44 & 24.66 (+2.55) & 44.82M \\
 & ST + ASR \& MT   & 21.58  & 26.37 & 26.17& 24.71 (+2.60)  & 56.81M\\
 & ST + MAM           & 20.78  & 25.34 & 24.46 & 23.53 (+1.42)& 33.15M \\
 & ST + MAM + ASR   & 22.41  & 26.89 & 26.49 & 25.26 (+3.15) & 46.72M\\
 & \citet{liu2020bridging}  & 22.55  & - & -         & -    & - \\
 & \citet{le2020dual}      & 23.63  & 28.12 & 27.55 & 26.43 (+4.32) & 51.20M\\
 & Cascade$^{\S}$      & 23.65  & 28.68 & 27.91 & 26.75 (+4.64) & 83.79M \\
\cmidrule(l){2-7} 
 & FAT-ST (base).     & 22.70  & 27.86 & 27.03 & 25.86 (+3.75) & 39.34M \\
\cmidrule(l){1-7} 
\multirow{2}{*}{
        ASR \& MT 
}
  & ST                 & 21.95 & 26.83  & 26.03 & 24.94 (+2.83) & 31.25M \\
 & ST + ASR \& MT  & 22.05 & 26.95 & 26.15  & 25.05 (+2.94)     & 56.81M\\
\cmidrule(l){1-7} 
 MAM  & FAT-ST (base) &22.29 & 27.21 & 26.26 & 25.25 (+3.14) & 39.34M \\
\cmidrule(l){1-7} 
\multirow{2}{*}{
        FAT-MLM  
}
 & FAT-ST (base) & \textbf{23.68} &  28.61 & \textbf{27.84} & 26.71 (+4.60) & 39.34M \\
 & FAT-ST (big) & 23.64  & \textbf{29.00} & 27.64 & \textbf{26.76} (+4.65)  & 58.25M\\

\midrule
\end{tabular}
}
\caption{
BLEU comparisons on Must-C test set between our proposed methods and other baselines over 3 translation directions using 
MuST-C ($\dsxy$) only 
(including pretraining methods).
$^{\S}$ are reported in \citet{inaguma2020espnet}.}
\label{tb:multiresults}
\end{table*}

\begin{table*}[ht]
\centering
\resizebox{1.9\columnwidth}{!}{
\begin{tabular}{@{}cc|c|lcccc@{}}
\toprule
Pretrain Data
& Pretrain Method & Train Data & Models & \multicolumn{1}{c}{En$\to$De} & \multicolumn{1}{c}{En$\to$Es} & \multicolumn{1}{c}{En$\to$Nl} & Avg. \\ \midrule

\multirow{2}{*}{
$\emptyset$
}
& 
\multirow{9}{*}{\begin{tabular}[c]{@{}c@{}}  \end{tabular}
}
&
\multirow{8}{*}{
\begin{tabular}[c]{@{}c@{}}$\dsxy$ \end{tabular} 
} 
 & ST                   & 19.64  & 23.68 & 23.01 & 22.11 \\
& & & Cascade$^{\S}$      & 23.65  & 28.68 & 27.91 & 26.75 (+4.64) \\
\cmidrule(l){1-2} 
\cmidrule(l){4-8} 
 \multirow{4}{*}{
    \begin{tabular}[c]{@{}c@{}}
        $\dsxy \cup \dsx \cup \dxy$
    \end{tabular} 
}
&
\multirow{2}{*}{
        ASR \& MT 
}
&
  & ST                 & 22.20  & 27.16  & 26.15 & 25.17 (+3.06) \\
& & & ST + ASR \& MT  & 22.73 & 27.99 & 27.12  & 25.95 (+3.84)\\
\cmidrule(l){2-2} 
\cmidrule(l){4-8} 
& 
\multirow{2}{*}{
        FAT-MLM 
}
&
   & FAT-ST (base) & 23.98 & 28.95 & 28.08 & 27.00 (+4.89) \\
& &  & FAT-ST (big)  & 24.34 & 29.41 & 28.86 & 27.54 (+5.43) \\
\cmidrule(l){1-2} 
\cmidrule(l){4-8} 
\multirow{2}{*}{
    \begin{tabular}[c]{@{}c@{}}
        $\dsxy \cup \dsx \cup \dxy$\\
        $\cup \ds \cup \dx \cup \dy$
    \end{tabular} 
}
& 
\multirow{2}{*}{
        FAT-MLM 
}
&
  & FAT-ST (base) & 24.02 & 29.25   & 28.28 & 27.18 (+5.07)  \\
& & & FAT-ST (big)  & 24.58 & 30.10   & 29.36 & 28.01 (+5.90)\\
\midrule

\multirow{2}{*}{
    \begin{tabular}[c]{@{}c@{}}
        $\dsxy \cup \dsx \cup \dxy$\\
        $\cup \ds \cup \dx \cup \dy$
    \end{tabular}
}
& 
\multirow{2}{*}{
        FAT-MLM 
}
&
\multirow{2}{*}{
    \begin{tabular}[c]{@{}c@{}}
 $\dsxy$ \\
 $\dsx \cup \dxy$
    \end{tabular} 
}
& 
FAT-ST (base) & 23.91 & 29.01 & 28.18 &  27.03 (+4.92) \\
& &
& FAT-ST (big)  & \textbf{25.47} & \textbf{30.75} & \textbf{30.08} & \textbf{28.77 (+6.66)} \\
\midrule
$\emptyset$ &   & $\dsxy$ + $\dsy'$
& \citet{pino2020self}  & 25.2 &  -  & -  & -  \\
\midrule
\end{tabular}
}
\caption{
BLEU comparisons on Must-C test set between our proposed methods 
using additional data. $\dsx$: Librispeech,
$\dxy$: Europarl MT, $\ds$: libri-light,
$\dx, \dy$: monolingual data from Europarl or Wiki Text.
$^{\S}$ are reported in \citet{inaguma2020espnet}.
\citet{pino2020self}
use extra $\dsy'$ which includes
Librispeech ($\dsx$) and 35,217 hour version of
Libri-light speech data (almost $10\times$ of our $\ds$)
paired with their corresponding 
pseudo-translations 
generated
by ASR and MT models.
Their model size is 435.0M.
}
\label{tb:ptresults}
\end{table*}

Raw audio files are processed by Kaldi~\cite{Povey11thekaldi}
to extract 80-dimensional log-Mel filterbanks stacked with 3-dimensional pitch features
using a window size of 25 ms  and step size of 10 ms.
We train sentencepiece~\cite{kudo-richardson-2018-sentencepiece} 
models with a joint vocabulary size of 8K for text in each dataset. 
Training samples that have more than 3000 frames have been ignored for GPU efficiency.
Our basic Transformer-based E2E-ST framework has similar settings 
with ESPnet-ST\cite{inaguma2020espnet}. 
the speech input is first down-sampled the speech input with 2 layers of 
2D convolution of size 3 with stride size of 2.
Then there is a standard 12-layers Transformer
with feed-forward layer of 2048 hidden size to
bridge the source and target side.
We only use 4 attention heads on each side of the transformer 
and each of them has a dimensionality of 256.
We also show the results of FAT-ST big model with 4096 hidden size for 
feed-forward layers of all transformer layer.
For speech reconstruction module, we simply linearly project the outputs of the Transformer
encoder to another latent space, then upsample the latent representation
with 2-layers deconvolution to match the size of the original input signal.
We choose 30\% for the random masking ratio $\lambda$ across all the 
experiments including pre-training. 
During inference, we do not perform any masking over the speech input.
We average the last 5 checkpoints for testing.
For decoding,
we use a beam search with beam-size 5
and length penalty 0.6 for German, 
0.0 for Spanish and 0.3 for Dutch.

\subsection{Translation Quality  Comparisons}

We showcase the translation accuracy of FAT-ST comparing against to
the baselines in Table~\ref{tb:multiresults}
and Table~\ref{tb:ptresults}:
\begin{itemize}
\item \textbf{ST}: this is the vanilla speech translation system which does not use transcriptions.
\vspace{-5pt}
\item \textbf{ST + ASR MTL}: ST model with an additional ASR decoder and is trained with ASR multi-task learning using the transcriptions.
\vspace{-5pt}
\item \textbf{ST + ASR \& MT MTL}: ST model with
an additional ASR decoder
and a MT encoder. It is trained with ASR and MT multi-task learning.
\vspace{-5pt}
\item \textbf{ST + MAM}: ST trained with additional MAM loss~\cite{chen2020mam} which can be formalized
as $\ell_{\vecs}(\ds)$ (See Fig.~\ref{fig:mam}).
\vspace{-5pt}
\item \textbf{ST + MAM + ASR MTL}: ST trained with MAM loss and ASR multi-task learning.
\vspace{-5pt}
\item \textbf{\citet{liu2020bridging}}: 
An end-to-end ST system with a multimodal encoder. 
\vspace{-5pt}
\item \textbf{\citet{le2020dual}}:
The state-of-the-art end-to-end ST model with an extra ASR decoder.
\item \textbf{Cascade}: cascaded model which first transcribes the speech into
transcription then passes the results to a machines translation system.
\item \textbf{ST + ASR \& MT pretraining}: the encoder of ST is initialized by a 
pretrained ASR encoder and decoder initialized by a pretrained MT decoder
\item \textbf{\citet{pino2020self}}:
They propose to leverage additional speech data
by generating pseudo-translations using a cascaded
or an end-to-end speech translation model.
\end{itemize}


\begin{table}[tb!]
\centering
\resizebox{.6\columnwidth}{!}{
\begin{tabular}{l c}
\toprule
Model          & \# Parameters \\ \hline
  MAM & 23.69 M \\ 
  FAT-MLM (base) & 25.76 M \\ 
  FAT-MLM (big)  & 38.36 M \\ 
\midrule
\end{tabular}
}
\caption{
Models sizes of different models. 
}
\label{tb:model_size}
\end{table}

\subsubsection{Model Size of Pretraining Models}

Table~\ref{tb:model_size} shows the number of parameters 
of different pretraining models.
We can see that our FAT-MLM base model is a little bit larger than the MAM
pretraining model, and the FAT-MLM big model is much larger than the
base model.

\subsubsection{Training with $\dsxy$}

In Table \ref{tb:multiresults},
with no pretraining, we can see that our proposed
FAT-ST base model achieves the best results
except \citet{le2020dual} and the cascaded model. 
However, our base model has much less
 parameters than both of them.
Models with ASR or MT MTL and \citet{liu2020bridging}
all use the transcription data in Must-C dataset
but show worse performance,
thus our model can use transcription data more
efficiently.
Similar to other open source ST implementation results on Must-C
\footnote{\hyperlink{https://github.com/espnet/espnet}{ESPnet: https://github.com/espnet/espnet}},
our implementation of ST + ASR \& MT MTL is worse
than ST + ASR.

We also compare the performance of models pretrained
from different pretraining models.
With pretrained on Must-C, 
FAT-ST (base) is improved by 0.85 BLEU by being finetuned from FAT-MLM,
while it's performance drops by finetuning from MAM.
Meanwhile, our proposed methods achieve much better performance
compared with ASR \& MT pretraining baselines.
We also note that our FAT-ST base model
for the first time achieves
similar performances compared with Cascade baselines in
these three translation directions of Must-C, 
while comparing with the cascaded model,
our our base model is much smaller in size
and faster in inference (see Fig.~\ref{fig:speed}).

\subsubsection{Pretraining with Additional Data}

Table \ref{tb:ptresults} shows that
FAT-MLM can further improve FAT-ST by simply
adding speech recognition data $\dsx$ (Librispeech)
text machine translation data $\dxy$ (Europarl) and
even speech only data $\ds$ (Libri-light) and
monolingual text data $\dx \cup \dy$.
This shows good representation learning ability
of our proposed FAT-MLM models.
We can see that using larger data, the performance
of our big model is increased much faster than
the base model.
That's because the number of parameters of the base model
is too limited to learn from such big data.

\subsubsection{Finetuning with Additional Data}

The last part of Table~\ref{tb:multiresults}
show that FAT-ST can be improved by learning from extra
speech recognition and machine translation data.
This is promising because speech translation data
is very limited compared with much more abundant 
speech recognition and machine translation data.
Different from \citet{pino2020self}
who propose to leverage additional speech data
by generating pseudo-translations, our method
doesn't use any pseudo-labels.
Our best model
outperforms their result on En$\to$De by using 
much $7\times$ smaller model size and almost $10\times$
smaller speech data.

\begin{table}[t!]
\centering
\resizebox{1.0\columnwidth}{!}{
\begin{tabular}{@{}ccllccc@{}}
\toprule
Train Data & Pretrain Data & Models & \multicolumn{1}{c}{$\to$De} & \multicolumn{1}{c}{$\to$Es} & \multicolumn{1}{c}{$\to$Nl} \\ \midrule
\multirow{10}{*}{
\begin{tabular}[c]{@{}c@{}}$\dsxy$ \end{tabular} 
}
&
\multirow{2}{*}{
\begin{tabular}[c]{@{}c@{}}No \\ pretraining \end{tabular} 
}

& MT$^{\S}$  &  27.63  & 32.61 & 32.08 \\
\cmidrule(l){3-6} 
& & FAT-ST (base) & 24.41 & 30.81 & 29.18    \\

\cmidrule(l){2-6} 

& 
\multirow{2}{*}{
    \begin{tabular}[c]{@{}c@{}}
         $\dsxy$
    \end{tabular} 
}
  & FAT-ST (base) & 27.24 & 31.98 & 31.27 \\
\cmidrule(l){3-6}
&  & FAT-ST (big)  & 26.92 & 32.29 & 31.48 \\
\cmidrule(l){2-6} 
& 
\multirow{2}{*}{
    \begin{tabular}[c]{@{}c@{}}
        $\dsxy$ \\
        $\cup \dsx \cup \dxy$
    \end{tabular} 
}
   & FAT-ST (base) & 27.43 & 32.38 & 32.44  \\
\cmidrule(l){3-6} 
&  & FAT-ST (big)  & 27.60  & 32.95 & 32.37  \\
\cmidrule(l){2-6} 
& 
\multirow{2}{*}{
    \begin{tabular}[c]{@{}c@{}}
        $\dsxy\cup \dsx \cup \dxy$ \\
        $ \cup \ds \cup \dx \cup \dy$
    \end{tabular} 
}
  & FAT-ST (base) & 27.63 & 32.75 & 32.52 \\
\cmidrule(l){3-6} 
& & FAT-ST (big) & 28.13  & 33.39  & 32.72 \\

 \midrule
\multirow{2}{*}{
    \begin{tabular}[c]{@{}c@{}}
 $\dsxy$  \\ $\cup \dsx \cup \dxy$
    \end{tabular} 
}
&
\multirow{2}{*}{
    \begin{tabular}[c]{@{}c@{}}
        $\dsxy\cup \dsx \cup \dxy$ \\
        $ \cup \ds \cup \dx \cup \dy$
    \end{tabular} 
}
& FAT-ST (base)  & 27.89 & 32.96  & 32.43 \\
& & FAT-ST (big) & 28.80 & 34.28  & 34.22 \\
\midrule
\end{tabular}
}
\caption{
Comparisons of the auxiliary MT task between MT baselines and our proposed methods.
$^{\S}$ are reported in \citet{inaguma2020espnet}.}
\label{tb:mt}
\end{table}

\begin{table}[ht!]
\centering
\resizebox{.75\columnwidth}{!}{
\begin{tabular}{l|c}
\toprule
Model           & En$\to$De \\ \hline
FAT-ST with FAT-MLM (base)         & 23.68    \\ \hline
\qquad \qquad - FAT-MLM decoder init.     &  23.20    \\
\qquad \qquad - FAT-MLM encoder init.     & 22.70      \\
\qquad \qquad - CTC loss     & 22.30      \\
\qquad \qquad - Hierarchical Transformer & 22.07      \\ 
\qquad \qquad - FAT-MLM loss & 20.64      \\ 
\qquad \qquad - MT loss      & 19.64      \\ 
\midrule
\end{tabular}
}
\caption{Ablation study. Here, hierarchical transformer
means the model only shares the 6 layers of the transformer encoder
for acoustic feature input and text feature input.}
\label{tb:ablation}
\end{table}

\begin{table*}[t!]
\centering
\resizebox{1.3\columnwidth}{!}{
\begin{tabular}{@{}ll@{}}
\toprule
Speech transcription   &  those are their expectations of who you are not yours      \\
\cmidrule(l){2-2} 
Target reference  & 那 \,  是   他们  所期望的    你的 \; 样子 \;\; 而不是 你自己的 期望   \\
& {\small \it       that \, is \, they \; expected \;\; your appearance \: not \;\;\;\;  yourself \; expectation } \\
\midrule
Cascade-ASR   &      those are \textbf{\textcolor{red}{there}} expectations \textbf{\textcolor{red}{to do}} you are not yours   \\
\cmidrule(l){2-2} 
Cascade-Translation &  那些     都是   希望   \;  \textbf{\textcolor{red}{做到的}}  , \;\;\; \textbf{\textcolor{red}{你    不是  你的 }} 。 \\
& {\small \it          those \: are \; expect  achievement \; you  not \;\;  yours  }  \\
\midrule
FAT-ST  &     这些    是 \, 他们    对    你的 \;\; 期 望 , \; 而不是 你的 \;\; 期望 。      \\
& {\small \it these \: are \; they \; to \: your expectation \;\; not \;\;\;\; your \;\; expectation } \\
\midrule
\end{tabular}
}
\caption{
English-to-Chinese speech translation example. 
The cascaded system is our implementation using the TED training data.
The errors of cascaded model is highlighted in red.
}
\label{tb:example}
\end{table*}

\subsubsection{Performance of Auxiliary MT Task}

Table~\ref{tb:mt} shows the translation quality of
auxiliary MT task of FAT-ST.
Although our models trained with Must-C
are worse than the MT baseline, 
by using FAT-MLM trained with more data,
our proposed methods can easily outperform
the MT baseline.
Note that 
these models' parameters are tuned to
optimize speech translation task
and MT is just an auxiliary task.

\subsubsection{Ablation Study}

Table~\ref{tb:ablation} shows an ablation study of
our proposed method.
we can see that all the components contribute to
the final performance.

\begin{table}[t!]
\centering
\resizebox{.6\columnwidth}{!}{
\begin{tabular}{@{}lc@{}}
\toprule
Models & \multicolumn{1}{c}{En$\to$Zh}  \\ \midrule
KD \cite{liu2019end}    & 19.55     \\
LUT \cite{dong2020listen}     &  20.84  \\
COSTT \cite{dong2021consecutive}   &  21.12   \\
Cascade \cite{dong2020listen}      &  21.36  \\
ST*  & 22.07     \\
\cmidrule(l){1-2} 
FAT-ST  & 23.73     \\
FAT-MLM + FAT-ST  & 25.49     \\
\midrule
\end{tabular}
}
\caption{
BLEU comparisons on English-to-Chinese speech translation. 
* is our implementation. 
Cascaded model is implemented by \citet{dong2020listen}.
}
\label{tb:enzh}
\end{table}

\subsubsection{English$\to$Chinese Speech Translation}

We also compare several models in TED English$\to$Chinese
speech translation task \cite{liu2019end} with 524 hours speech
in training set, 1.5 hours validation set (dev2010) 
and 2.5 hours test set (tst2015).
We follow our previous experiments to preprocess the data.
Same with previous work, we evaluate the performance with character-level BLEU.
Table \ref{tb:enzh} shows that our proposed model can largely outperform other baselines.
Table \ref{tb:example} shows one example in this dataset.
The translation of the cascaded model is wrong because of the
errors in the its ASR (their$\to$their, of who$\to$ to do),
while our FAT-ST produces the right translation.

\begin{figure}[t!]
\centering
\includegraphics[width=6cm]{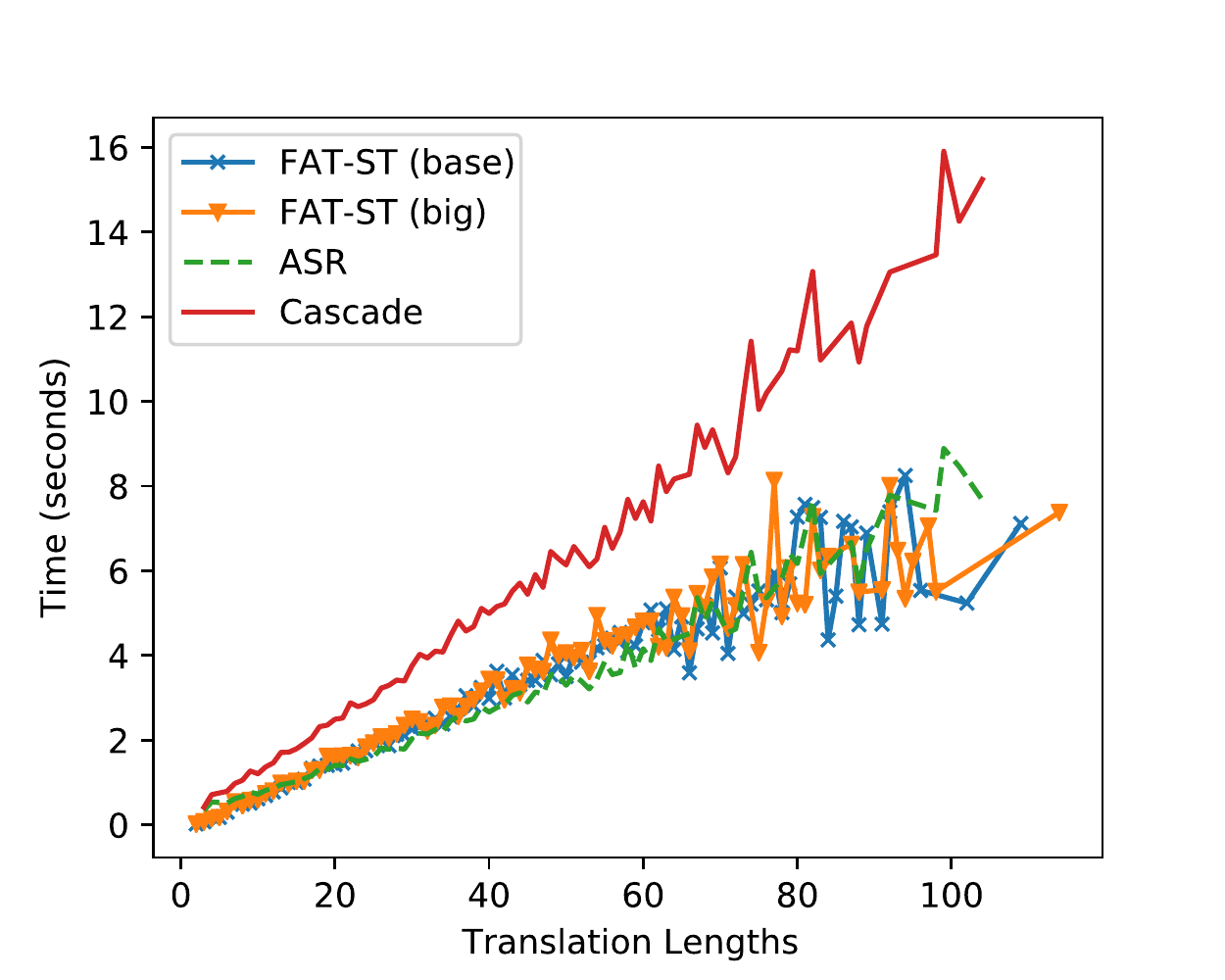}
\caption{Decoding time comparison between Cascaded model (including its ASR) and FAT-ST.}
\vspace{-1.0cm}
\label{fig:speed}
\end{figure}

\subsubsection{Decoding Speed}

Fig.~\ref{fig:speed} shows 
the decoding speed comparison between the Cascade model
and our proposed FAT-ST.
Our proposed FAT-ST model is almost $2\times$ faster than 
the Cascade system which needs to wait for
the speech recognition module to finish before starting to translate.
The decoding time of FAT-ST (big) is almost the
same as FAT-ST (base) because we only increase the
feedforward network in Transformers.

\section*{Conclusion}

In this paper, we propose Fused Acoustic and Text
Masked Language Model (FAT-MLM) which learns
a unified representation for text and speech
from any data that combines speech and text.
We further extend this framework to
a sequence-to-sequence
speech translation model which
enables learning from speech recognition
and text-based machine translation data at the first time.
Our results show significant improvement
on three translation directions of the Must-C
dataset and outperform the cascaded baseline.

\section*{Acknowledgements}

We thank Kenneth Church and Jiahong Yuan for discussions,
and Juneki Hong for proofreading, 
and the anonymous reviewers for suggestions.




\bibliography{main}
\bibliographystyle{icml2021}


%



\end{CJK}

\end{document}